\documentclass[a4paper, 10pt, conference]{IEEEtran}      

\usepackage{graphicx}
\usepackage{tcolorbox}
\usepackage{url} 
\usepackage{array}
\usepackage{booktabs}
\usepackage{comment}
\usepackage{balance}
\usepackage{array}
\usepackage{multirow}

\newcommand\MyBox[2]{
  \fbox{\lower0.75cm
    \vbox to 1.7cm{\vfil
      \hbox to 1.7cm{\hfil\parbox{1.4cm}{#1\\#2}\hfil}
      \vfil}%
  }%
}

\IEEEoverridecommandlockouts                              

\title{\LARGE \bf
Beyond Vision: How Large Language Models Interpret Facial Expressions from Valence-Arousal Values
}

\author{\parbox{16cm}{\centering
    {\large Vaibhav Mehra$^{1,*}$, Guy Laban$^{2,*}$, and Hatice Gunes$^2$}\\
    {\normalsize
    $^1$ HEI-Lab, Universidade Lusófona, Lisbon, Portugal\\
    $^2$ Department of Computer Science and Technology, University of Cambridge, Cambridge, United Kingdom}}
    \thanks{*Vaibhav Mehra and Guy Laban have contributed equally to this work and share first authorship. Corresponding author: Guy Laban - Guy.Laban@cl.cam.ac.uk.}
}

\begin{document}

\thispagestyle{empty}
\pagestyle{empty}
\pagestyle{plain}
\maketitle

\begin{abstract}

Large Language Models primarily operate through text-based inputs and outputs, yet human emotion is communicated through both verbal and non-verbal cues, including facial expressions. While Vision-Language Models analyze facial expressions from images, they are resource-intensive and may depend more on linguistic priors than visual understanding. To address this, this study investigates whether LLMs can infer affective meaning from dimensions of facial expressions—Valence-Arousal values, structured numerical representations, rather than using raw visual input. VA values were extracted using Facechannel from images of facial expressions and provided to LLMs in two tasks: (1) categorizing facial expressions into basic (on the IIMI dataset) and complex emotions (on the Emotic dataset) and (2) generating semantic descriptions of facial expressions (on the Emotic dataset). Results from the categorization task indicate that LLMs struggle to classify VA values into discrete emotion categories, particularly for emotions beyond basic polarities (e.g., happiness, sadness). However, in the semantic description task, LLMs produced textual descriptions that align closely with human-generated interpretations, demonstrating a stronger capacity for free-text affective inference of facial expressions. 


\end{abstract}


\section{Introduction}

Large Language Models (LLMs) are predominantly text-based models, designed to process and generate human language naturally. When used as interactive agents, these models rely heavily on verbal inputs and outputs to infer and express emotions. However, human emotion extends beyond words; non-verbal cues, such as facial expressions, convey crucial affective meanings essential to communication \cite{he2017memd, toisoul2021estimation}. Affective communication with artificial agents should include both verbal and non-verbal cues \cite{Share2024} and is influenced by users' emotional states \cite{Open2023}, underscoring the need for AI systems that can interpret and generate meaningful representations of human emotions. As LLMs are increasingly employed in applications requiring emotional intelligence \cite{lbot04}, it is vital to assess their ability to move beyond merely language processing. Accordingly, this work examines the extent to which LLMs can understand and interpret facial expressions, addressing the gap between verbal and non-verbal affective communication in intelligent system design.

Vision-Language Models (VLMs) are used to analyse facial expressions by processing raw visual inputs. These models extract information from images and videos to infer emotions \cite{liu2024speak,yang2024emollm}. However, relying on raw image processing presents several challenges: it requires significant computational resources and raises privacy concerns in sensitive contexts. Moreover, while VLMs demonstrate strong performance in emotion recognition \cite{lei2024large, yao2024vlm}, their reliance on visual input remains unclear. Many of these models integrate multimodal data, but their outputs may be predominantly shaped by linguistic priors rather than genuine visual understanding \cite{Lin2023,Luo2024}. This lack of transparency makes it difficult to assess the role of visual information in their predictions. An alternative approach is to represent emotional information in a structured format, rather than relying on raw visual inputs. One such representation is Valence-Arousal (VA) values \cite{barrett1998discrete}, which quantify expressions along two dimensions: Valence (positivity/negativity) and Arousal (intensity). If models can interpret facial expressions effectively using only VA values, this would reduce reliance on direct image processing while maintaining emotional interpretability. 


This approach allows us to assess whether LLMs can generalize affective meaning from structured numerical representations of facial expressions, rather than relying on explicit image features. 
Therefore, this study explores whether LLMs’ semantic reasoning can be extended beyond language to structured affective data, offering insights into their latent capacity for cross-modal inference. 
Specifically, we evaluate their ability in two key tasks: \textbf{(1)} categorizing facial expressions into discrete emotional labels and \textbf{(2)} generating semantic descriptions of these expressions. By comparing LLM-generated outputs to human annotations, this study provides insights into the strengths and limitations of LLMs in non-verbal emotion recognition. To achieve these objectives, the study addresses the following research questions:  
\begin{itemize}  
\item [\textbf{RQ1}] \textit{To what extent LLMs can predict basic emotion categories from facial expressions using only VA values?  }
\item [\textbf{RQ2}] \textit{To what extent LLMs can semantically describe facial expressions from VA values, and how closely do these descriptions align with human-annotated descriptions?  }
\end{itemize}

Accordingly, the potential contributions of this study are: 

\begin{itemize}
\item Investigating LLMs’ ability to interpret facial expressions from VA values rather than direct visual input, assessing whether structured numerical representations are sufficient for affective inference.

\item Evaluating the semantic coherence of LLM-generated descriptions by comparing them to human annotations, providing insights into how LLMs conceptualize and verbalize expressions.

\item Examining whether categorical classification or semantic description is more effective for LLMs in deriving meaningful interpretations from VA values, providing insights into their strengths and limitations in structured versus free-form affective inference of facial expressions.


\end{itemize}





\section{Method}

In this study, we conducted two experiments with two distinct datasets. The IIMI dataset \cite{TEWARI_Mehta_Srinivasan_2023} and the Emotic dataset \cite{kosti2017emotic} as detailed in Sections \ref{iimi} and \ref{emotic}. Each dataset included images of facial expressions that were processed using FaceChannel \cite{barros2020facechannel}, an off the shelf package that predicts VA values ranging from -1 to 1 and categorizes facial expressions into basic emotional categories. The extracted VA values were input into LLMs through custom prompts to classify these into categories of emotion (in \textbf{Experiment 1}) or describe the expressions semantically (in \textbf{Experiment 2}). Accordingly, the outputs were analyzed to (\textbf{Experiment 1}) show LLMs' ability to classify expressions from VA values into emotional categories, and (\textbf{Experiment 2}) to demonstrate the extent of similarity between textual descriptions of facial expressions generated by LLMs (based on VA values) to those given by humans (based on their observation). 

\section{Experiment 1: Categorization task}


Humans often describe facial expressions via variety of different categories of emotion \cite{Du2014,Ekman1992,Jack2016}. To understand LLMs' ability to classify VA values to categories of emotions, a categorization experiment was conducted. The experiment included two sub-experiments. \textbf{Experiment 1.1} tested LLMs' ability to classify VA values into basic emotions (see \cite{ekman1992there}). 
Considering that expressions can correspond to a complex range of emotions, where an expression may align with multiple categories \cite{Liu2022}, \textbf{Experiment 1.2} evaluated  LLMs' ability of mapping VA values also to complex emotions (see \cite{burkitt2002complex}) via a multi-class categorisation task with a larger dataset. 

\subsection{Experiment 1.1: Basic Emotion Categorization}
\label{exp1.1}

\subsubsection{IIMI dataset}
\label{iimi}

The IIMI dataset \cite{TEWARI_Mehta_Srinivasan_2023} contains 700 images of Indian individuals expressing seven basic emotions defined by Ekman’s model (see \cite{ekman1992there}). 
The dataset includes 100 images per category, each assigned to a single emotion, making it ideal for single-class classification tasks  \cite{liu2024speak}. 

\subsubsection{Methodology}

All images from the IIMI dataset \cite{TEWARI_Mehta_Srinivasan_2023}, were processed with the two models of FaceChannel \cite{barros2020facechannel}. The categorization model classified images into basic emotion categories: Neutral, Happiness, Surprise, Sadness, Anger, Disgust, Fear, and Contempt, consistent with the IIMI dataset. The dimensional model extracted VA values, which were input into the LLM model, GPT-4o-mini \cite{gpt40mini}, using the following prompt:

\begin{tcolorbox}
\small\textit{``The value of valence is [valence\_value], the arousal value is [arousal\_value]. Categorize the associated facial expression in one of the following categories: anger, disgust, fear, happiness, sadness, surprise, or neutral. Respond in no more than a single category."}
\end{tcolorbox}


\subsubsection{Analysis}

Accuracy was calculated as the proportion of correctly categorised images relative to the total number of images in the dataset. The accuracy values for both models were compared to evaluate their performance in emotion classification
\subsubsection{Results}
\label{exp1.1_res}

Both models 
performed poorly, with accuracies of 30.42\% and 31.42\%, respectively, and show bias towards specific emotion categories. GPT 4o-mini achieves near perfect accuracy for Happiness (87\%) and Sadness (98\%), 22\% for Fear, and almost none for other categories. FaceChannel perfectly predicts Sad and Neutral, achieves 20\% accuracy for Happiness but fails for the rest (See Table \ref{tab:conf_mat_combined}).

\begin{table*}
  \caption{Confusion Matrices for GPT-4o-mini and FaceChannel Predictions}
  \label{tab:conf_mat_combined}
  \centering
  \resizebox{\textwidth}{!}{%
  \begin{tabular}{l | c c c c c c c | c c c c c c c}
    \toprule
    & \multicolumn{7}{c|}{\textbf{GPT-4o-mini}} & \multicolumn{7}{c}{\textbf{FaceChannel}} \\
    \textbf{Category} & \textbf{Happy} & \textbf{Fear} & \textbf{Neutral} & \textbf{Surprise} & \textbf{Disgust} & \textbf{Sad} & \textbf{Angry} 
    & \textbf{Happy} & \textbf{Fear} & \textbf{Neutral} & \textbf{Surprise} & \textbf{Disgust} & \textbf{Sad} & \textbf{Angry} \\
    \midrule
    Happy   &  87 &  0  & 13  &  0  &  0  &  0  &  0  &  20  &  0  & 80  &  0  &  0  &  0  &  0  \\
    Fear    &  0  & 22  &  0  &  0  &  0  & 78  &  0  &  0   &  0  & 34  &  0  &  0  & 66  &  0  \\
    Neutral & 56  &  0  &  5  &  0  &  0  & 39  &  0  &  0   &  0  & 100 &  0  &  0  &  0  &  0  \\
    Surprise& 93  &  0  &  0  &  1  &  0  &  0  &  0  &  0   &  0  & 100 &  0  &  0  &  0  &  0  \\
    Disgust &  0  &  0  &  0  &  0  &  0  & 100 &  0  &  0   &  0  & 89  &  0  &  0  & 11  &  0  \\
    Sad     &  2  &  0  &  0  &  0  &  0  & 98  &  0  &  0   &  0  &  0  &  0  &  0  & 100 &  0  \\
    Angry   & 39  &  0  &  0  &  0  &  0  & 61  &  0  &  0   &  0  &  0  &  0  &  0  & 100 &  0  \\
    \bottomrule
  \end{tabular}%
  }
\end{table*}

\subsection{Experiment 1.2: Complex Emotion Categorization}
\label{exp1.2}



\subsubsection{Emotic Dataset}
\label{emotic}
The Emotic dataset \cite{kosti2017emotic} includes diverse scenarios with individual faces, multiple faces, and social situations. The dataset consists of 12,821 images in the training subset and 3,663 images in the test subset. 
Since the study includes an evaluation task rather than a training task, we used the test subset, which includes 3,047 images with clear facial expressions (after manual inspection). This sample provided sufficient power for statistical analysis (\(\alpha = .05\), 1 - \(\beta = .8\), \(d = .2\)) while also minimizing computational costs and environmental impact \cite{faiz2023llmcarbon}.
Each image in the dataset is annotated with 1 to 9 categories of emotion (according to \cite{kosti2017emotic}) out of 26 categories, ranging from basic \cite{ekman1992there} to more complex emotions \cite{burkitt2002complex,Liu2022}. Each image in the dataset is annotated with VA values by humans while following the method of Mou et al. \cite{mou2015group}. The Emotic dataset was ideal for the task as it includes diverse facial expressions and multi-class emotion labels, enabling an evaluation of LLMs' multi-class categorization abilities in this affective domain.

\subsubsection{Methodology}

The images from the Emotic dataset \cite{kosti2017emotic} were processed using FaceChannel's  dimensional model \cite{barros2020facechannel} extracting VA values for each image, which were then provided to the LLMs using the following prompt:

\begin{tcolorbox}
\small\textit{``The value of valence is [valence\_value], the arousal value is [arousal\_value]. Classify the image into [n\_categories] of the most relevant categories from the following 26: Peace, Affection, Esteem, Anticipation, Engagement, Confidence, Happiness, Pleasure, Excitement, Surprise, Sympathy, Doubt/Confusion, Disconnection, Fatigue, Embarrassment, Yearning, Disapproval, Aversion, Annoyance, Anger, Sensitivity, Sadness, Disquietment, Fear, Pain, and Suffering. Respond only with comma-separated category."}
\end{tcolorbox}

Two LLM models, GPT-4o-mini \cite{gpt40mini} and GPT-4o \cite{gpt40}, classified each unit in the dataset based on its VA values (both those provided in the dataset, as well as those extracted using FaceChannel) into \textit{n} emotion categories, corresponding to the number of human-annotated categories. This led to a total of 10,633 classifications for the 3,047 images in the dataset. We utilised GPT-4o-mini due to its lower cost and reduced environmental impact. However, given its poor performance in Experiment 1.1 and the complexity of the task, we also tested GPT-4o.
\subsubsection{Analysis}
Two metrics were calculated to evaluate the multi-class classification task: the percentage of images where at least one predicted category matched the human annotations and the percentage where all predicted categories were an exact match.

\subsubsection{Results}

GPT 4o-mini correctly predicted at least one category for 49.67\% of images and all categories for 18.32\% of the images. With FaceChannel VA values, it achieved 50.75\% for at least one correct category and 11.01\% for all categories. GPT 4o, using FaceChannel VA values, had lower results: 43.26\% for at least one correct category and 6.91\% for all categories. Surprisingly, GPT-4o performed worse than GPT-4o-mini. The poor accuracy across all cases suggests that LLMs struggle with complex or overlapping emotions beyond basic polarised emotions such as happiness or sadness. Conventional machine learning models may handle such nuanced tasks more effectively \cite{xenos2024vllms}.

\section{Experiment 2: Semantic Description task}
\label{exp2}

LLMs perform better at generating semantically descriptive outputs compared to outputs that are syntactically correct but lack meaningful semantic content \cite{lee2024learning}. This is because their primary use case has been language generation, and they are trained accordingly. Moreover, facial expressions do not always align with discrete emotion categories, as the same expression can convey different emotions and social information \cite{Barrett2019}. As a result, describing expressions in words—capturing their intensity, subtlety, and affective dimensions—may provide a more accurate and flexible representation than rigid classification into predefined emotion labels \cite{Lecker2021, Li2022}. In addition, comparing AI-generated affective explanations to human explanations provides insight into how well AI systems align with human reasoning and social norms \cite{Dogan2025}. Thus, to address the limitations of Experiment 1, \textbf{Experiment 2} aimed at evaluating LLMs' performance in semantically describing facial expressions using only VA values extracted from images.


\subsection{Methodology} 

We used the FaceChannel dimensional model \cite{barros2020facechannel} to extract VA values from 3,047 images in the test subset (see Section \ref{emotic}) of the EMOTIC dataset \cite{kosti2017emotic}. The Emotic dataset was ideal for the task as it includes diverse facial expressions with human-annotated explanations, enabling a comparison to LLMs' semantic descriptions. These were then submitted to the LLMs to generate \textit{n} semantic descriptions for each unit in the dataset, corresponding to the number of human-annotated descriptions of facial expressions, using the following prompt:


\begin{tcolorbox}
\small\textit{``The value of valence is [valence\_value], the arousal value is [arousal\_value]. What do you understand from these about the emotions expressed by the facial expressions. In only [n\_categories] independent sentences, describe the expressed emotion and mental states, without mentioning the valence and arousal values."}
\end{tcolorbox}

This led to a total of 10,633 semantic descriptions for the 3,047 images. Three LLMs were tested: GPT 4o, GPT 4o-mini, and LLAMA 3.2 8B Instruct. LLAMA 3.2, an open-source model, offers better accessibility for future research and for replicating the paradigm. This comparison aimed at evaluating performance and generalization. 

\subsection{Analysis}

Semantic similarity between the original and LLM-generated descriptions was calculated using two methods and three models. The first method, \textit{combined semantic similarity}, compares the full LLM-generated description with the concatenated definitions of all human-assigned categories. The second method, \textit{separate semantic similarity}, treats each LLM sentence and category definition independently, calculates an \textit{n × n} similarity matrix, and averages the values. Vector representations of sentences were created using Transformers \cite{vaswani2017attention}, Word2Vec \cite{church2017word2vec}, and BERT \cite{kenton2019bert}. Word2Vec represents words as dense vectors based on co-occurrence patterns in a corpus, capturing local semantic relationships but lacking contextual awareness. In contrast, Transformer-based models dynamically adjust word embeddings based on surrounding context, allowing for a deeper understanding of sentence structure and meaning. BERT, specifically, leverages bidirectional context, making it particularly effective at capturing nuanced semantic relationships \cite{Apidianaki2023}. Cosine similarity was used to compute the final scores.

\begin{table*}

  \caption{Examples of semantically similar GPT-4o-mini predicted descriptions with semantic similarity.} 
  \label{tab:4}
  \renewcommand{\arraystretch}{1.5} 
  \centering
  \resizebox{\textwidth}{!}{%
  \begin{tabular}{>{\centering\arraybackslash}m{3cm}
                  >{\centering\arraybackslash}m{1cm}
                  >{\centering\arraybackslash}m{1cm}
                  >{\centering\arraybackslash}m{3cm} 
                  >{\centering\arraybackslash}m{3cm} 
                  >{\centering\arraybackslash}m{3cm} 
                  >{\centering\arraybackslash}m{3cm} 
                  >{\centering\arraybackslash}m{3cm}}
    \toprule
   \normalsize \textbf{Image} & \normalsize \textbf{Valence} & \normalsize \textbf{Arousal} & \normalsize \textbf{Human Description} & \normalsize \textbf{GPT-4o-mini Description} & \multicolumn{3}{c}{\normalsize \textbf{Semantic similarity}} \\
    \cmidrule(lr){6-8}
    & & & & & \normalsize \textbf{Word2Vec} & \normalsize \textbf{Transformers} & \normalsize \textbf{BERT} \\
    \midrule
    \includegraphics[width=2cm]{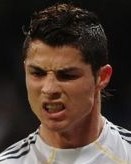} & \normalsize -.39 & \normalsize	.61 &  \normalsize \textbf{Disapproval}: feeling that \textbf{something is wrong} or reprehensible; contempt; hostile & \normalsize A state of \textbf{anxiety} or \textbf{agitation}, individual feels \textbf{unease} but is also \textbf{alert} and activated. & \normalsize	 84.27\% & \normalsize	 50.67\% & \normalsize	 76.08\% \\
    \includegraphics[width=2cm]{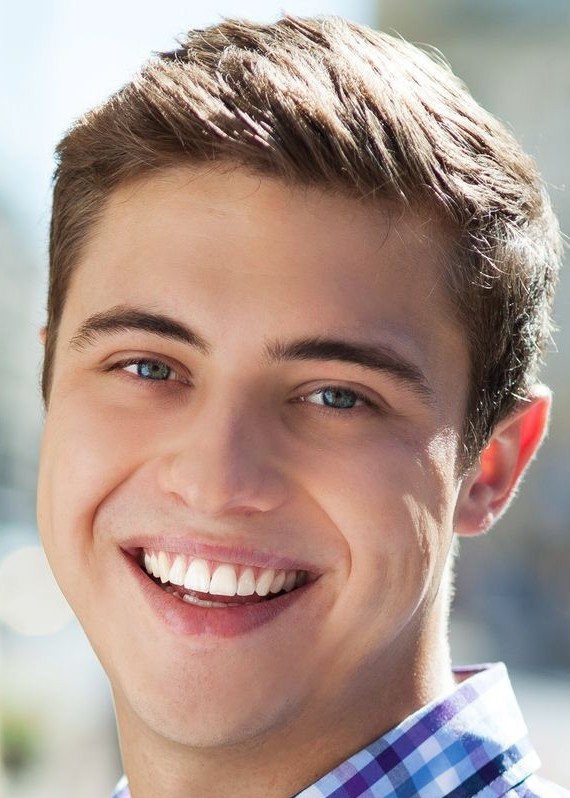} & \normalsize .88 & \normalsize -.08 &  
 \normalsize	\textbf{Happiness}: Feeling \textbf{delighted}, feeling \textbf{enjoyment} or amusement & \normalsize A moderately \textbf{positive emotional state} characterized by mild \textbf{enthusiasm} or \textbf{contentment}, suggesting a sense of \textbf{optimism} or satisfaction without overwhelming excitement. & \normalsize 82.93\% & \normalsize 55.06\% & \normalsize	78.68\% \\
    \bottomrule
  \end{tabular}%
  }
\end{table*}

To assess the generalizability of the cosine similarity results, Bootstrap testing calculated average cosine similarity scores and 95\% confidence intervals. Since violations of normality are not a concern with large samples like the one used in our study due to the Central Limit Theorem \cite{Altman1995,Knief2021}, and non-parametric tests may be too sensitive with such large samples \cite{Fagerland2012}, a one-sample \textit{t}-test to determine whether similarity scores exceeded the baseline score of .5, which represents similarity above random chance \cite{corley2005measuring}.

\subsection{Results}
\label{exp_2_res}

For the \textbf{GPT-4o-mini}, applying the Word2Vec model with the combined method of similarity calculation yielded an average cosine similarity of $M = .81$, 95\%CI [.81, .82]. A one-sample t-test confirmed that this mean similarity was significantly higher than the baseline value of .5, $t(3046) = 262.57, p < .001$. Using the separate method of similarity calculation for the same model resulted in an average cosine similarity of $M = .72$, 95\%CI [.72, .73], also significantly higher than the baseline, $t(3046) = 259.84, p < .001$. When using the Transformer-based embeddings, the combined method produced a lower similarity of $M = .42$, 95\%CI [.42, .43], and a one-sample t-test indicated that this result was not significantly different from the baseline value, $t(3046) = -40.06, p = 1$. The separate method with Transformer embeddings yielded $M = .31$, 95\%CI [.31, .32], $t(3046) = -31.6, p = 1$. With BERT-based embeddings, the combined method showed an average similarity of $M = .79$, 95\%CI [.78, .79], $t(3046) = 555.14, p < .001$, while the separate method resulted in $M = .62$, 95\%CI [.62, .63], $t(3046) = 182.94, p < .001$.

For the \textbf{GPT-4o}, Word2Vec embeddings with the combined method yielded an average similarity of $M = .80$, 95\%CI [.80, .81], significantly higher than the baseline ($t(3046) = 227.54, p < .001$). The separate method resulted in $M = .74$, 95\%CI [.73, .74], $t(3046) = 225.68, p < .001$. Using Transformer embeddings, the combined method resulted in $M = .39$, 95\%CI [.39, .40], $t(3046)=-60.23, p = 1$, while the separate method produced $M=.28$, 95\%CI [.28, .29], $t(3046)=-67.84, p = 1$. For BERT embeddings, the combined method produced $M=.79$, 95\%CI [.79, .80], $t(3046)=520.12, p < .001$, while the separate method resulted in $M=0.62$, 95\%CI [.61, .62], $t(3046)=473.61, p < .001$.

For the \textbf{LLAMA 3.2 8B Instruct}, Word2Vec embeddings with the combined method produced an average similarity of $M=.77$, 95\%CI [.76, .77], $t(3046)=174.54, p < .001$. The separate method resulted in $M=.75$, 95\%CI [.74, .75], $t(3046)=173.28, p < .001$. For Transformer embeddings, the combined method produced $M=.35$, 95\%CI [.34, .35], $t(3046)=-80.49, p = 1$, while the separate method resulted in $M=.32$, 95\%CI [.32, .33], $t(3046) = -2.2, p = .98$. With BERT embeddings, the combined method showed $M=.75$, 95\%CI [.75, .76], $t(3046)=285.61, p < .001$, while the separate method resulted in $M=.66$, 95\%CI [.65, .66], $t(3046)=239.87, p < .001$. See Table \ref{tab:3} for the results and Table \ref{tab:4} for examples comparing human-annotated descriptions to those generated by the LLMs.

\begin{table}

\caption{Bootstrap mean results and \textit{t}-test results for Cosine similarity results of Experiment 2.}
\label{tab:3}
\resizebox{\columnwidth}{!}{%
\begin{tabular}{lccc}
    \toprule
    \textbf{Test} & \textbf{Word2Vec} & \textbf{Transformers} & \textbf{BERT} \\
    \midrule
    GPT-4o-mini (Combine) & .81$^{***}$ [.81, .82] & .42 [.42, .43] & .79$^{***}$ [.78, .79] \\
    GPT-4o-mini (Separate) & .72$^{***}$ [.72, .73] & .31 [.31, .32] & .62$^{***}$ [.62, .63] \\
    GPT-4o (Combine) & .80$^{***}$ [.80, .81] & .39 [.39, .40] & .79$^{***}$ [.79, .80] \\
    GPT-4o (Separate) & .74$^{***}$ [.73, .74] & .28 [.28, .29] & .62$^{***}$ [.61, .62] \\
    LLAMA (Combine) & .77$^{***}$ [.76, .77] & .35 [.34, .35] & .75$^{***}$ [.75, .76] \\
    LLAMA (Separate) & .75$^{***}$ [.74, .75] & .32 [.32, .33] & .66$^{***}$ [.65, .66] \\
    \midrule
    \multicolumn{4}{l}{\footnotesize Note: \( p < 0.001 = *** \)} \\
    \bottomrule
\end{tabular}%
}
\end{table}

\section{Discussion}

Our findings highlight both the potential and limitations of LLMs in inferring facial expressions from VA values alone. 
In Experiment 1, LLMs struggled to map VA values to discrete emotions. 
Biases were evident, with better performance for polarized emotions (e.g., happiness, sadness) but poor recognition of others (e.g., anger, surprise). Multi-class categorization of complex emotions improved performance slightly, yet exact matches were low, suggesting difficulty in capturing nuanced and complex emotions. In contrast, Experiment 2 showed that LLMs perform significantly better when generating open-ended semantic descriptions of facial expressions. This aligns with prior research indicating LLMs excel in free-text generation over rigid classification \cite{Xu2024,Sherburn2024}. BERT and Word2Vec performed better than Transformers, suggesting that pre-trained embeddings capturing contextual and semantic relationships are more effective than purely structural representations for mapping VA values to meaningful affective descriptions, highlighting the importance of leveraging linguistic priors when using LLMs for structured affective inference tasks.

The stronger performance in the semantic description task suggests that LLMs are more effective at inferring general affective meanings from VA values rather than rigidly categorizing them. This aligns with theories of emotion and affect, which posit that affective perception is often more gradient-based than categorical \cite{Barrett2016, Gendron2009, Satpute2016}, also when observing facial expressions \cite{Fujimura2011}. Our findings also highlight the potential for LLMs to complement multimodal emotion recognition systems by providing descriptive information rather than binary classifications. However, LLMs’ reliance on linguistic priors may lead to oversimplifications. Future work should explore integrating additional context (e.g., speech, action units) and comparing LLMs with VLMs to enhance emotion recognition.

\section{Conclusions}

In this study, we explored the ability of LLMs to classify and describe facial expressions based solely on VA values, shedding light on their potential for affective inference without direct visual input. LLMs performed notably better at generating semantic descriptions of expressions than at categorising emotions, indicating their strength in free-text descriptions over rigid classification tasks. These findings suggest that LLMs can process structured affective data, yet their reliance on linguistic priors may limit their ability to fully capture nuanced emotions. Overall, LLMs show promise for affective computing and facial expression research (e.g., see \cite{Zhang2024}), but require further refinement for nuanced emotional understanding. Hybrid approaches combining structured affective data with multimodal inputs could improve robustness. Future work should integrate multimodal inputs and refine LLMs' affective reasoning capabilities to enhance their application in privacy-conscious emotion recognition and social interactions. 




\section*{ETHICAL IMPACT STATEMENT}

This study did not involve human participants or personal data collection. This study utilized publicly available datasets, ensuring compliance with ethical standards for data use. By leveraging structured data rather than raw visual inputs, this research contributes to the advancement of privacy-conscious approaches in emotion recognition. The methods employed promote ethical AI development by reducing reliance on personally identifiable data and mitigating potential biases associated with direct human observation. A key ethical consideration is the potential for LLM-generated interpretations of affective data to reflect linguistic biases present in their training data. Future research should ensure that models are evaluated across diverse datasets to enhance fairness and generalizability in affective computing applications. Another consideration is the interpretability of LLMs' affective inferences. While this study examines their ability to describe emotions based on structured data, these models may not fully capture the complexity of human affective states. Over-reliance on LLM-generated interpretations in sensitive applications, such as mental health, should be approached with caution to avoid misleading conclusions.

\section*{ACKNOWLEDGMENTS}
V. Mehra is funded by the European Union Erasmus Mundus Joint Master Grant no: 101048710. G. Laban and H. Gunes are supported by the EPSRC project ARoEQ under grant ref. EP/R030782/1.

{\small
\bibliographystyle{ieee}
\balance{\bibliography{egbib}}
}

\end{document}